\documentclass{article}

\usepackage{graphicx}
\usepackage{subfigure}

\usepackage{natbib}
\setcitestyle{round}

\usepackage{algorithm}
\usepackage{algorithmic}

\usepackage[colorlinks,citecolor=blue,urlcolor=blue]{hyperref}

\usepackage{amsmath}
\usepackage{amssymb}
\usepackage{todonotes}
\usepackage{url}

\usepackage[margin=1.2in]{geometry}

\newcommand{\given}{\,|\,}
\newcommand{\pmin}{p_{\textrm{min}}}
\newcommand{\x}{\mathbf{x}}
\newcommand{\EI}{\textrm{EI}}
\newcommand{\concon}{\mathcal{C}}
\newcommand{\E}{\mathbb{E}}

\usepackage{authblk}

\title{Bayesian Optimization with Unknown Constraints}
\author{Michael A. Gelbart}
\author{Jasper Snoek}
\author{Ryan P. Adams}
\affil{School of Engineering and Applied Sciences, Harvard University}
\affil{\{mgelbart, jsnoek, rpa\} @ seas.harvard.edu}

\date{}

\begin{document} 

\maketitle

\begin{abstract} 
Recent work on Bayesian optimization has shown its effectiveness in global optimization of difficult black-box objective functions.  Many real-world optimization problems of interest also have constraints which are unknown \emph{a priori}.  In this paper, we study Bayesian optimization for constrained problems in the general case that noise may be present in the constraint functions, and the objective and constraints may be evaluated independently. We provide motivating practical examples, and present a general framework to solve such problems.  We demonstrate the effectiveness of our approach on optimizing the performance of online latent Dirichlet allocation subject to topic sparsity constraints, tuning a neural network given test-time memory constraints, and optimizing Hamiltonian Monte Carlo to achieve maximal effectiveness in a fixed time, subject to passing standard convergence diagnostics.
\end{abstract}

\section{Introduction}
\label{section:intro}

Bayesian optimization~\citep{Mockus1978} is a method for performing global optimization of unknown ``black box'' objectives that is particularly appropriate when objective function evaluations are expensive (in any sense, such as time or money). For example, consider a food company trying to design a low-calorie variant of a popular cookie. In this case, the design space is the space of possible recipes and might include several key parameters such as quantities of various ingredients and baking times. Each evaluation of a recipe entails computing (or perhaps actually measuring) the number of calories in the proposed cookie. Bayesian optimization can be used to propose new candidate recipes such that good results are found with few evaluations. 

Now suppose the company also wants to ensure the \emph{taste} of the cookie is not compromised when calories are reduced. Therefore, for each proposed low-calorie recipe, they perform a taste test with sample customers. Because different people, or the same people at different times, have differing opinions about the taste of cookies, the company decides to require that at least 95\% of test subjects must like the new cookie.
This is a constrained optimization problem: 
\[  \min_\x \, c(\x) \; \mathrm{s.t.} \; \rho(\x) \geq 1-\epsilon \;, \] 
where $\x$ is a real-valued vector representing a recipe, $c(\x)$ is the number of calories in recipe $\x$, $\rho(\x)$ is the fraction of test subjects that like recipe $\x$, and ${1-\epsilon}$ is the minimum acceptable fraction, i.e., 95\%. 

This paper presents a general formulation of constrained Bayesian optimization that is suitable for a large class of problems such as this one. Other examples might include tuning speech recognition performance on a smart phone such that the user's speech is transcribed within some acceptable time limit, or minimizing the cost of materials for a new bridge, subject to the constraint that all safety margins are met. 

Another use of constraints arises when the search space is known \emph{a priori} but occupies a complicated volume that cannot be expressed as simple coordinate-wise bounds on the search variables. For example, in a chemical synthesis experiment, it may be known that certain combinations of reagents cause an explosion to occur. This constraint is not unknown in the sense of being a discovered property of the environment as in the examples above---we do not want to discover the constraint boundary by trial and error explosions of our laboratory. Rather, we would like to specify this constraint using a boolean noise-free oracle function that declares input vectors as valid or invalid. Our formulation of constrained Bayesian optimization naturally encapsulates such constraints.

\subsection{Bayesian Optimization}

Bayesian optimization proceeds by iteratively developing a global statistical model of the unknown objective function. Starting with a prior over functions and a likelihood, at each iteration a posterior distribution is computed by conditioning on the previous evaluations of the objective function, treating them as observations in a Bayesian nonlinear regression. An \emph{acquisition function} is used to map beliefs about the objective function to a measure of how promising each location in input space is, if it were to be evaluated next. The goal is then to find the input that maximizes the acquisition function, and submit it for function evaluation.

Maximizing the acquisition function is ideally a relatively easy proxy optimization problem: evaluations of the acquisition function are often inexpensive, do not require the objective to be queried, and may have gradient information available.  Under the assumption that evaluating the objective function is expensive, the time spent computing the best next evaluation via this inner optimization problem is well spent. Once a new result is obtained, the model is updated, the acquisition function is recomputed, and a new input is chosen for evaluation. This completes one iteration of the Bayesian optimization loop.

For an in-depth discussion of Bayesian optimization, see \citet{brochu-etal-2010a} or \citet{lizotte-thesis}.  Recent work has extended Bayesian optimization to multiple tasks and objectives~\citep{krause-ong-2011,swersky-etal-2013a,Zuluaga:13} and high dimensional problems~\citep{wang-etal-2013,djolonga-2013}. Strong theoretical results have also been developed~\citep{Srinivas2010,Bull2011,defreitas-etal-2013a}.  Bayesian optimization has been shown to be a powerful method for the meta-optimization of machine learning algorithms~\citep{snoek-etal-2012b,BergstraJ2011} and algorithm configuration~\citep{hutter-2011a}.

\subsection{Expected Improvement}
\label{sub:ei}

An acquisition function for Bayesian optimization should address the exploitation vs.\ exploration tradeoff: the idea that we are interested both in regions where the model believes the objective function is low (``exploitation'') and regions where uncertainty is high (``exploration''). One such choice is the Expected Improvement (EI) criterion \citep{Mockus1978}, an acquisition function shown to have strong theoretical guarantees~\citep{Bull2011} and empirical effectiveness \citep[e.g.,][]{snoek-etal-2012b}. The expected improvement,~EI$(\x)$, is defined as the expected amount of improvement over some target $t$, if we were to evaluate the objective function at $\x$:
\begin{align}
\label{ei-definition}
\mathrm{EI}(\x) &= \mathbb{E}[(t-y)_+] = \int_{-\infty}^\infty (t-y)_+ p(y\given \x) \,\mathrm{d}y\;,
\end{align}
where $p(y\given\x)$ is the predictive marginal density of the objective function at $\x$, and ${(t-y)_+\equiv\max(0,t-y)}$ is the improvement (in the case of minimization) over the target~$t$. 
EI encourages both exploitation and exploration because it is large for inputs with a low predictive mean (exploitation) and/or a high predictive variance (exploration). Often,~$t$ is set to be the minimum over previous observations \citep[e.g.,][]{snoek-etal-2012b}, or the minimum of the expected value of the objective \citep{Brochu2010}. Following our formulation of the problem, we use the minimum expected value of the objective such that the probabilistic constraints are satisfied (see Section \ref{section:prob-con}, Eq., \ref{main-opt}). 

When the predictive distribution under the model is Gaussian, EI has a closed-form expression \citep{Jones2001}:
\begin{align}
\mathrm{EI}(\x) &= \sigma(\x) \left( z(\x)\Phi\left( z(\x)\right)+\phi\left( z(\x)\right) \right)
\end{align}
where ${z(\x) \equiv \frac{t-\mu(\x)}{\sigma(\x)}}$, $\mu(\x)$ is the predictive mean at~$\x$,~$\sigma^2(\x)$ is the predictive variance at $\x$, $\Phi(\cdot)$ is the standard normal CDF, and $\phi(\cdot)$ is the standard normal PDF. This function is differentiable and fast to compute, and can therefore be maximized with a standard gradient-based optimizer. In Section \ref{section:acq} we present an acquisition function for constrained Bayesian optimization based on EI.

\subsection{Our Contributions}
\label{sec:our_contributions}
The main contribution of this paper is a general formulation for constrained Bayesian optimization, along with an acquisition function that enables efficient optimization of such problems. Our formulation is suitable for addressing a large class of constrained problems, including those considered in previous work. The specific improvements are enumerated below. 

First, our formulation allows the user to manage uncertainty when constraint observations are noisy. 
By reformulating the problem with probabilistic constraints, the user can directly address this uncertainty by specifying the required confidence that constraints are satisfied.

Second, we consider the class of problems for which the objective function and constraint function need not be evaluated jointly. In the cookie example, the number of calories might be predicted very cheaply with a simple calculation, while evaluating the taste is a large undertaking requiring human trials. Previous methods, which assume joint evaluations, might query a particular recipe only to discover that the objective (calorie) function for that recipe is highly unfavorable. The resources spent simultaneously evaluating the constraint (taste) function would then be very poorly spent. We present an acquisition function for such problems, which incorporates this user-specified cost information.

Third, our framework, which supports an arbitrary number of constraints, provides an expressive language for specifying arbitrarily complicated restrictions on the parameter search spaces. For example if the total memory usage of a neural network must be within some bound, this restriction could be encoded as a separate, noise-free constraint with very low cost.
As described above, evaluating this low-cost constraint would take priority over the more expensive constraints and/or objective function.

\subsection{Prior Work} 

There has been some previous work on constrained Bayesian optimization.  \citet{gramacy2010} propose an acquisition function called the integrated expected conditional improvement (IECI), defined as 
\begin{align}
\mathrm{IECI}(\x) &= \int_{\mathcal{X}} \left[\mathrm{EI}(\x') - \mathrm{EI}(\x'|\x) \right]h(\x') \mathrm{d}\x'
\end{align}
In the above, EI$(\x')$ is the expected improvement at~$\x'$,~EI$(\x'|\x)$ is the expected improvement at $\x'$ given that the objective has been observed at $\x$ (but without making any assumptions about the observed value), and $h(\x')$ is an arbitrary density over $\x'$. In words, the IECI at $\x$ is the expected reduction in EI at $\x'$, under the density $h(\x')$, caused by observing the objective at $\x$. 
\citeauthor{gramacy2010} use IECI for constrained Bayesian optimization by setting~$h(\x')$ to the probability of satisfying the constraint. This formulation encourages evaluations that inform the model in places that are likely to satisfy the constraint.

\citet{Zuluaga:13} propose the Pareto Active Learning (PAL) method for finding Pareto-optimal solutions when multiple objective functions are present and the input space is a discrete set. Their algorithm classifies each design candidate as either Pareto-optimal or not, and proceeds iteratively until all inputs are classified. The user may specify a confidence parameter determining the tradeoff between the number of function evaluations and prediction accuracy. Constrained optimization can be considered a special case of multi-objective optimization in which the user's utility function for the ``constraint objectives'' is an infinite step function: constant over the feasible region and negative infinity elsewhere. However, PAL solves different problems than those we intend to solve, because it is limited to discrete sets and aims to classify each point in the set versus finding a single optimal solution.

\citet{snoek-2013a} discusses constrained Bayesian optimization for cases in which constraint violations arise from a failure mode of the objective function, such as a simulation crashing or failing to terminate. The thesis introduces the weighted expected improvement acquisition function, namely expected improvement weighted by the predictive probability that the constraint is satisfied at that input.

\subsection{Formalizing the Problem} 
\label{section:prob-con}

In Bayesian optimization, the objective and constraint functions are in general unknown for two reasons. First, the functions have not been observed everywhere, and therefore we must interpolate or extrapolate their values to new inputs. Second, our observations may be noisy; even after multiple observations at the same input, the true function is not known. Accounting for this uncertainty is the role of the model, see Section \ref{section:model}.

However, before solving the problem, we must first define it. Returning to the cookie example, each taste test yields an estimate of $\rho(\x)$, the fraction of test subjects that like recipe $\x$. But uncertainty is always present, even after many measurements. Therefore, it is impossible to be certain that the constraint~${\rho(\x)\geq 1-\epsilon}$ is satisfied for any~$\x$. Likewise, the objective function can only be evaluated point-wise and, if noise is present, it may never be determined with certainty. 

This is a stochastic programming problem: namely, an optimization problem in which the objective and/or constraints contain uncertain quantities whose probability distributions are known or can be estimated \citep[see e.g.,][]{Shapiro2009}. A natural formulation of these problems is to minimize the objective function \emph{in expectation}, while satisfying the constraints \emph{with high probability}. The condition that the constraint be satisfied with high probability is called a \emph{probabilistic constraint}. This concept is formalized below.

Let $f(\x)$ represent the objective function. Let $\mathcal{C}(\x)$ represent the the \emph{constraint condition}, namely the boolean function indicating whether or not the constraint is satisfied for input $\x$. 
For example, in the cookie problem,~${\mathcal{C}(\x)\iff \rho(\x)\geq 1-\epsilon}$.
Then, our probabilistic constraint is 
\begin{align}
\Pr(\mathcal{C}(\x)) \geq 1- \delta\,,
\end{align}
for some user-specified minimum confidence ${1-\delta}$.

If $K$ constraints are present, for each constraint~${k\in(1,\ldots ,K)}$ we define $\concon_k(\x)$ to be the constraint condition for constraint $k$. Each constraint may also have its own tolerance $\delta_k$, so
 we have $K$ probabilistic constraints of the form
\begin{align}
\Pr(\concon_k(\x))\geq 1-\delta_k \;.
\end{align}

All $K$ probabilistic constraints must ultimately be satisfied at a solution to the optimization problem.\footnote{Note: this formulation is based on individual constraint satisfaction for all constraints. Another reasonable formulation requires the (joint) probability that \emph{all} constraints are satisfied to be above some single threshold.}

Given these definitions, a general class of constrained Bayesian optimization problems can be formulated as 
\begin{align}
\label{main-opt}
\min_\x \, \mathbb{E}[f(\x)] \;\, \mathrm{s.t.} \; \forall k \; \Pr(\concon_k(\x)) \geq 1-\delta_k \,.
\end{align}
The remainder of this paper proposes methods for solving problems in this class using Bayesian optimization. Two key ingredients are needed: a model of the objective and constraint functions (Section \ref{section:model}), and an acquisition function that determines which input $\x$ would be most beneficial to observe next (Section \ref{section:acq}).

\section{Modeling the Constraints} 
\label{section:model}

\subsection{Gaussian Processes}
\label{sub:gp}
We use Gaussian processes (GPs) to model both the objective function $f(\x)$ and the constraint functions.
A GP is a generalization of the multivariate normal distribution to arbitrary index sets, including infinite length vectors or functions, and is specified by its positive definite covariance kernel function $K(\x,\x')$. 
GPs allow us to condition on observed data and tractably compute the posterior distribution of the model for any finite number of query points. A consequence of this property is that the marginal distribution at any single point is univariate Gaussian with a known mean and variance.  
See \citet{Rasmussen2006} for an in-depth treatment of GPs for machine learning. 

We assume the objective and all constraints are independent and model them with independent GPs. Note that since the objective and constraints are all modeled independently, they need not all be modeled with GPs or even with the same types of models as each other. Any combination of models suffices, so long as each one represents its uncertainty about the true function values.

\subsection{The latent constraint function, $g(\x)$} 
\label{sub:modeling}
In order to model constraint conditions $\concon_k(\x)$, we introduce real-valued latent \emph{constraint functions} $g_k(\x)$ such that for each constraint $k$, the constraint condition $\concon_k(\x)$ is satisfied if and only if~${g_k(\x) \geq 0}$.\footnote{Any inequality constraint ${g(\x)\leq g_0}$ or ${g(\x)\geq g_1}$ can be represented this way by transforming to a new variable ${\hat{g}(\x) \equiv g_0-g(\x)\geq 0}$ or ${\hat{g}(\x) \equiv g(\x)-g_1\geq 0}$, respectively, so we set the right-hand side to zero without loss of generality.} 
Different observation models lead to different likelihoods on~$g(\x)$, as discussed below.
By computing the posterior distribution of $g_k(\x)$ for each constraint, we can then compute ${\Pr(\concon_k(\x))=\Pr(g_k(\x)\geq 0)}$ by simply evaluating the Gaussian CDF using the predictive marginal mean and variance of the GP at $\x$. 

Different constraints require different definitions of the constraint function $g(\x)$. When the nature of the problem permits constraint observations to be modeled with a Gaussian likelihood, the posterior distribution of $g(\x)$ can be computed in closed form. If not, approximations or sampling methods are needed \citep[see][p. 41-75]{Rasmussen2006}.  We discuss two examples below, one of each type, respectively.

\subsection{Example I: bounded running time}
\label{sub:boundedrunningtime}
Consider optimizing some property of a computer program such that its running time $\tau(\x)$ must not exceed some value $\tau_{\mathrm{max}}$.
Because $\tau(\x)$ is a measure of time, it is nonnegative for all $\x$ and thus not well-modeled by a GP prior. We therefore choose to model time in logarithmic units. In particular, we define~${g(\x)=\log \tau_{\mathrm{max}} - \log \tau}$, so that the condition ${g(\x)\geq 0}$ corresponds to our constraint condition~${\tau \leq \tau_{\mathrm{max}}}$, and place a GP prior on $g(\x)$. For every problem, this transformation implies a particular prior on the original variables; in this case, the implied prior on~$\tau(\x)$ is the log-normal distribution. In this problem we may also posit a Gaussian likelihood for observations of~$g(\x)$. This corresponds to the generative model that constraint observations are generated by some true latent function corrupted with i.i.d.\ Gaussian noise. As with the prior, this choice implies something about the original function~$\tau(\x)$, in this case a log-normal likelihood. The basis for these choices is their computational convenience. Given a Gaussian prior and likelihood, the posterior distribution is also Gaussian and can be computed in closed form using the standard GP predictive equations.

\subsection{Example II: modeling cookie tastiness}
Recall the cookie optimization, and let us assume that constraint observations arrive as a set of counts indicating the numbers of people who did and did not like the cookies. We call these \emph{binomial constraint observations}. Because these observations are discrete, they are not modeled well by a GP prior. Instead, we model the (unknown) binomial probability $\rho(\x)$ that a test subject likes cookie $\x$, which is linked to the observations through a binomial likelihood.\footnote{We use the notation $\rho(\x)$ both for the fraction of test subjects who like recipe $\x$ and for its generative interpretation as the probability that a subject likes recipe $\x$.} In Section \ref{section:prob-con}, we selected the constraint condition ${\rho(\x)\geq 1-\epsilon}$, where~${1-\epsilon}$ is the user-specified threshold representing the minimum allowable probability that a test subject likes the new cookie. Because ${\rho(\x)\in(0,1)}$ and~${g(\x)\in\mathbb{R}}$, we define ${g(\x) = s^{-1}(\rho(\x))}$,
where $s(\cdot)$ is a monotonically increasing sigmoid function mapping ${\mathbb{R}\rightarrow (0,1)}$ as in logistic or probit regression.\footnote{When the number of binomial trials is one, this model is called Gaussian Process Classification.} In our implementation, we use~${s(z)=\Phi(z)}$, the Gaussian CDF. 
The likelihood of~$g(\x)$ given the binomial observations is then the binomial likelihood composed with~$s^{-1}$. Because this likelihood is non-Gaussian, the posterior distribution cannot be computed in closed form, and therefore approximation or sampling methods are needed.

\subsection{Integrating out the GP hyperparameters}
Following \citet{snoek-etal-2012b}, we use the Mat\'{e}rn~5/2 kernel for the Gaussian process prior, which corresponds to the assumption that the function being modeled is twice differentiable. This kernel has~${D+1}$ hyperparameters in~$D$ dimensions: one characteristic length scale per dimension, and an overall amplitude. Again following \citet{snoek-etal-2012b}, we perform a fully-Bayesian treatment by integrating out these kernel hyperparameters with Markov chain Monte Carlo (MCMC) via slice sampling \citep{Neal00slicesampling}.

When the posterior distribution cannot be computed in closed form due to a non-Gaussian likelihood, we use elliptical slice sampling \citep{murray2010} to sample $g(\x)$. We also use the prior whitening procedure described in \citet{Murray-Adams-2010a} to avoid poor mixing due to the strong coupling of the latent values and the kernel hyperparameters.

\section{Acquisition Functions}
\label{section:acq} 

\subsection{Constraint weighted expected improvement}
Given the probabilistic constraints and the model for a particular problem, it remains to specify an acquisition function that leads to efficient optimization. 
Here, we present an acquisition function for constrained Bayesian optimization under the Expected Improvement (EI) criterion (Section \ref{sub:ei}). However, the general framework presented here does not depend on this specific choice and can be used in conjunction with any improvement criterion.

Because improvement is not possible when the constraint is violated, we can define an acquisition function for constrained Bayesian optimization by extending the expectation in Eq. \ref{ei-definition} to include the additional constraint uncertainty. This results in a constraint-weighted expected improvement criterion, $a(\x)$:
\begin{align}
\label{weighted-ei}
a(\x) & = \EI(\x) \Pr(\concon(\x))  \\
      & = \EI(\x) \prod_{k=1}^K \Pr(\concon_k(\x))
\end{align}
where
 the second line follows from the assumed independence of the constraints. 

Then, the full acquisition function $a(\x)$, after integrating out the GP hyperparameters, is given by
\begin{align*}
a(\x)=\int \mathrm{EI}(\x|\theta)p(\theta |\mathbf{D})p(\mathcal{C}(\x)|\x, \mathbf{D}', \omega)p(\omega | \mathbf{D}') \mathrm{d}\theta\mathrm{d}\omega,
\end{align*}
where $\theta$ is the set of GP hyperparameters for the objective function model, $\omega$ is the set of GP hyperparameters for the constraint model(s), ${\mathbf{D}=\{\x_n,y_n\}_{n=1}^N}$ are the previous objective function observations, and $\mathbf{D}'$ are the constraint function observations. 

\subsection{Finding the feasible region} 
The acquisition function given above is not defined when at least one probabilistic constraint is violated for all $\x$, because in this case the
EI target does not exist and therefore EI cannot be computed. 
In this case we take the acquisition function to include only the second factor, 
\begin{align}
\label{all-violate}
a(\x)= \prod_{k=1}^K \Pr(g_k(\x)\geq 0)
\end{align}
Intuitively, if the probabilistic constraint is violated everywhere, we ignore the objective function and try to satisfy the probabilistic constraint until it is satisfied somewhere. This acquisition function may also be used if no objective function exists, i.e., if the problem is just to search for any feasible input.
This feasibility search is purely exploitative: it searches where the probability of satisfying the constraints is highest. This is possible because the true probability of constraint satisfaction is either zero or one. Therefore, as the algorithm continues to probe a particular region, it will either discover that the region is feasible, or the probability will drop and it will move on to a more promising region.

\begin{figure*}[t!]
\centering

\subfigure[True objective]{
  \includegraphics[height=0.17\columnwidth]{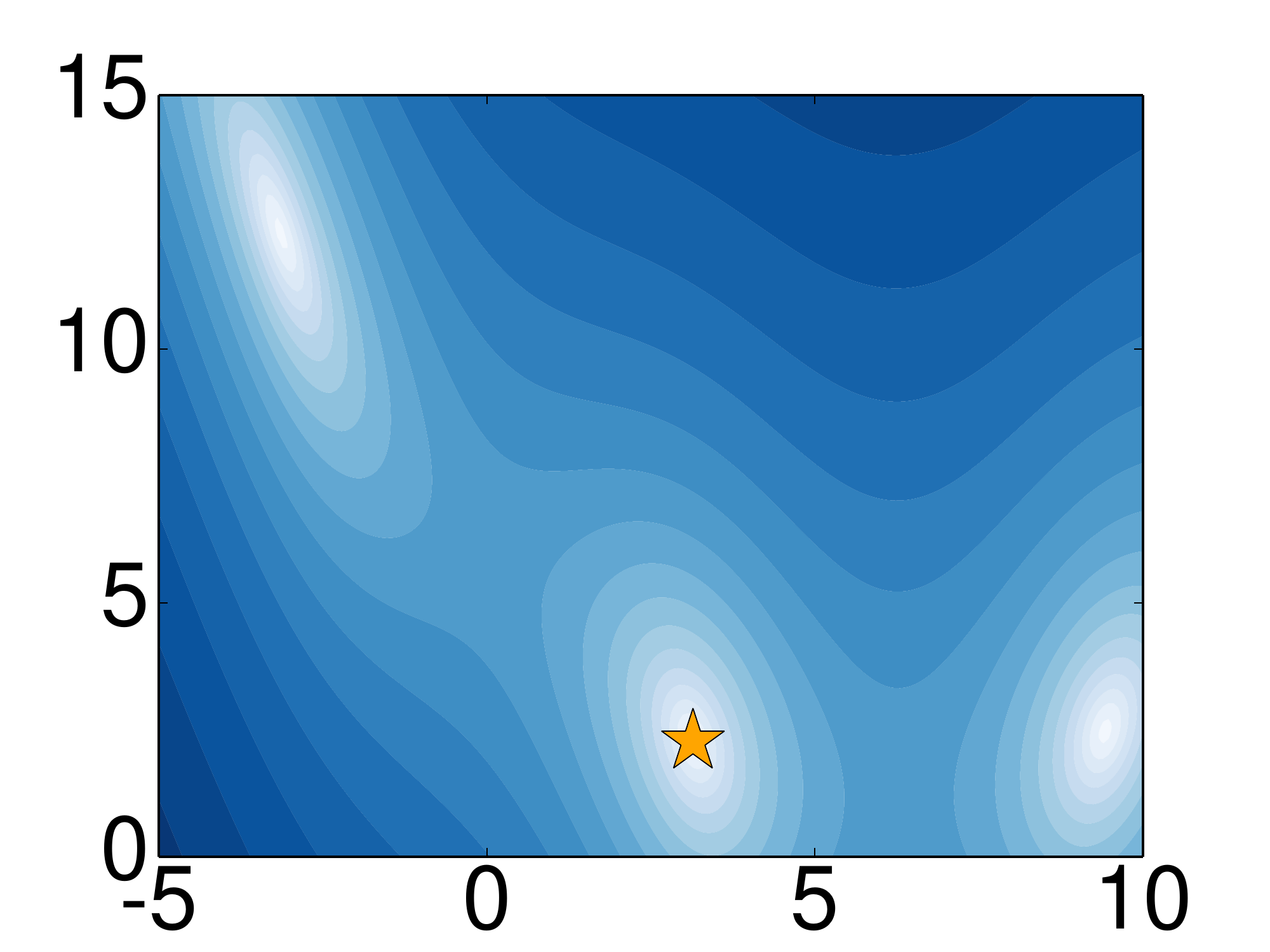}
  \label{fig:truebranin}
}
\subfigure[True constraint]{
  \includegraphics[height=0.17\columnwidth]{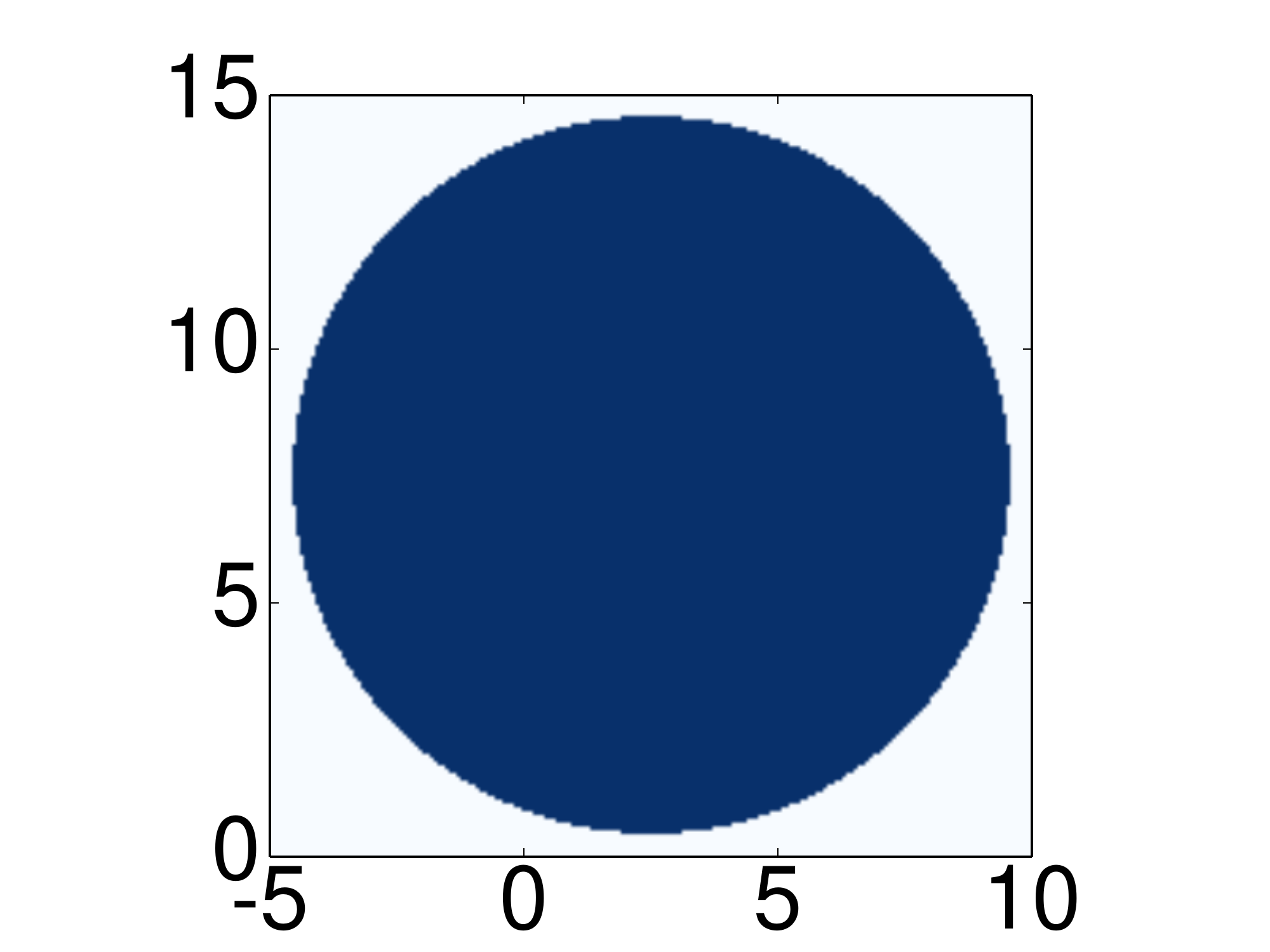}
  \label{fig:diskcon}
}
\subfigure[Objective GP mean]{
  \includegraphics[height=0.17\columnwidth]{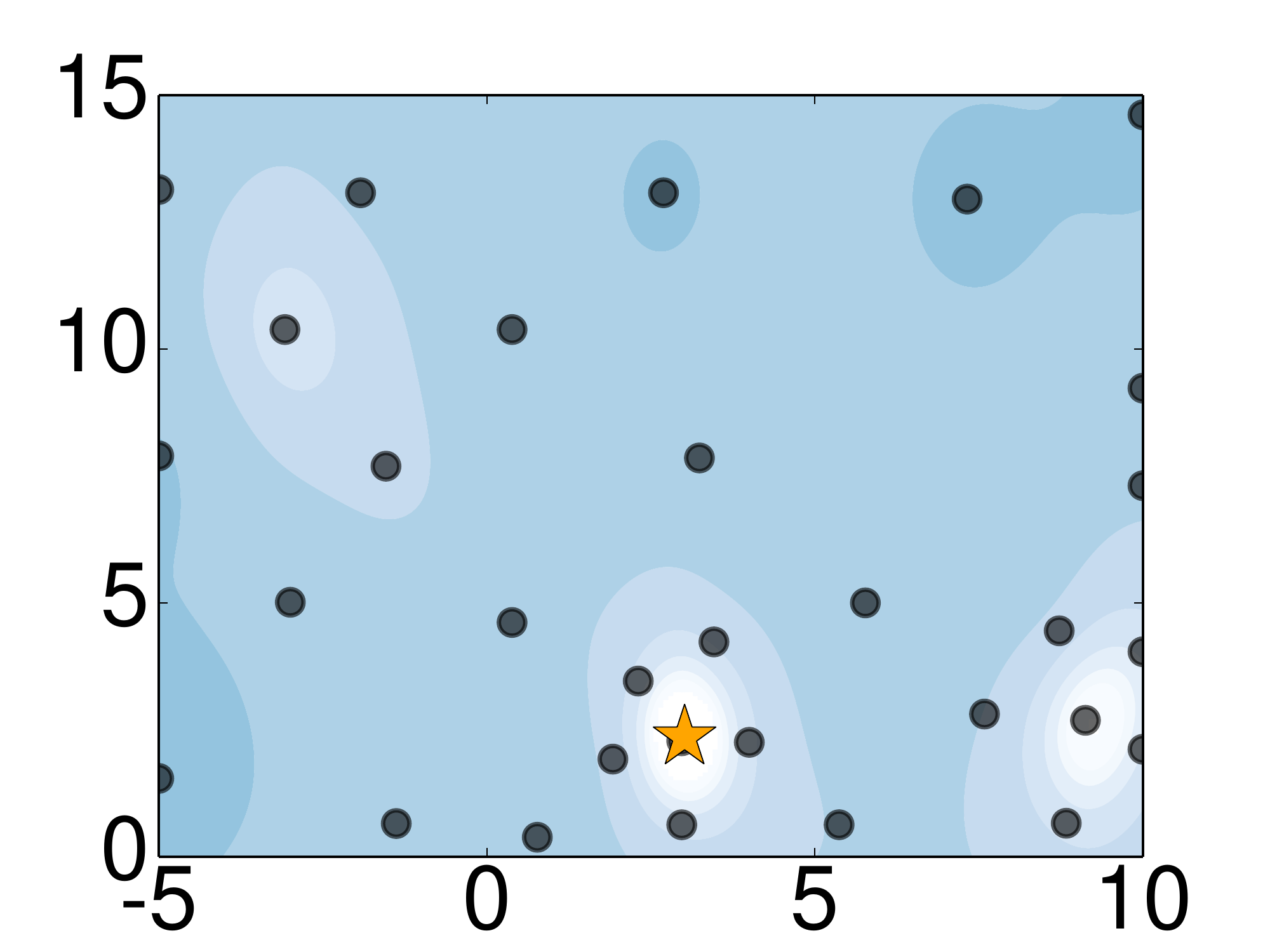}
  \label{fig:gpmean}
}
\subfigure[Objective GP variance]{
  \includegraphics[height=0.17\columnwidth]{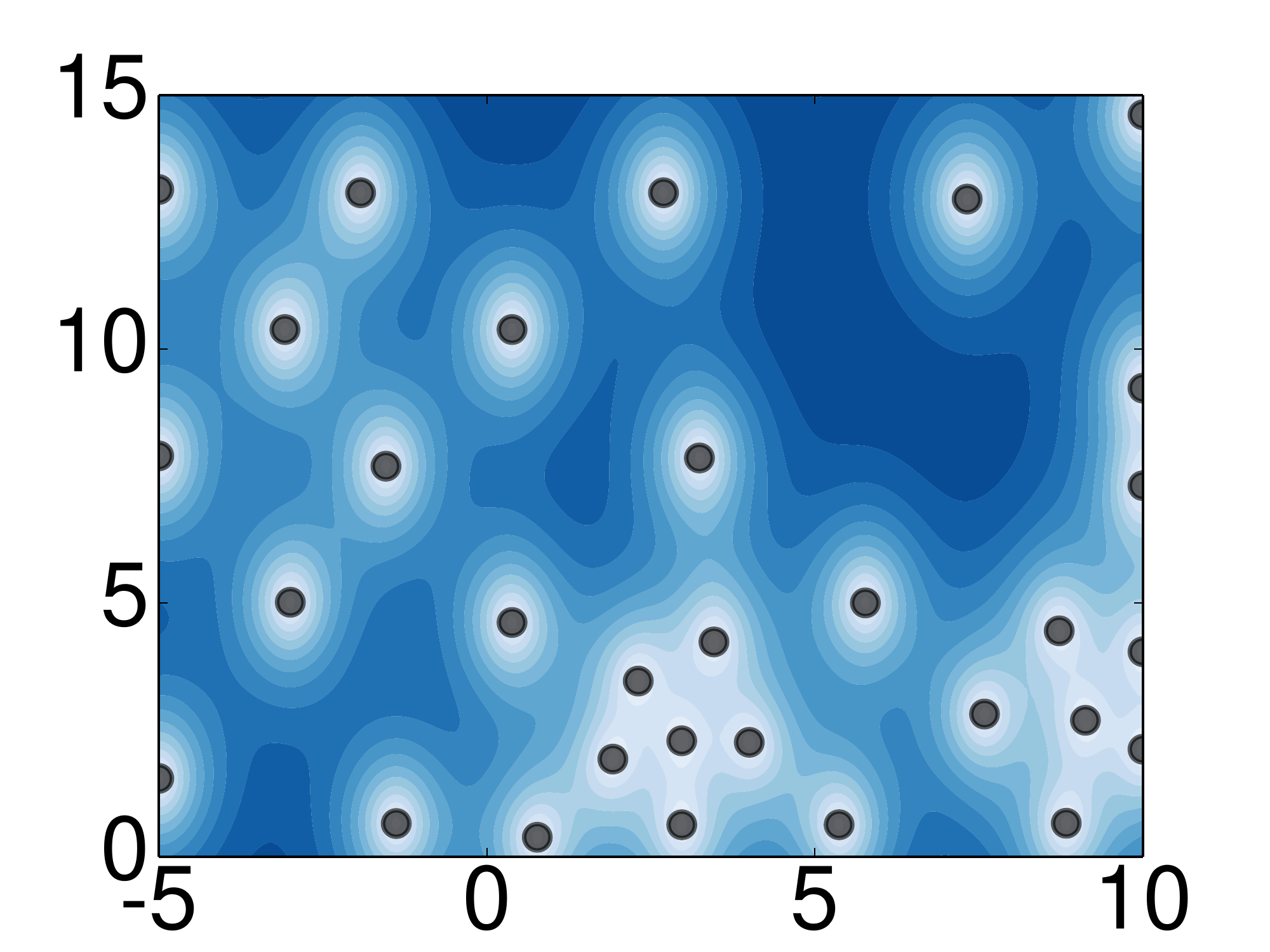}
  \label{fig:gpsigma}
}

\vspace{-1em}

\subfigure[$\Pr(g(\x)\geq 0)$]{
  \includegraphics[height=0.17\columnwidth]{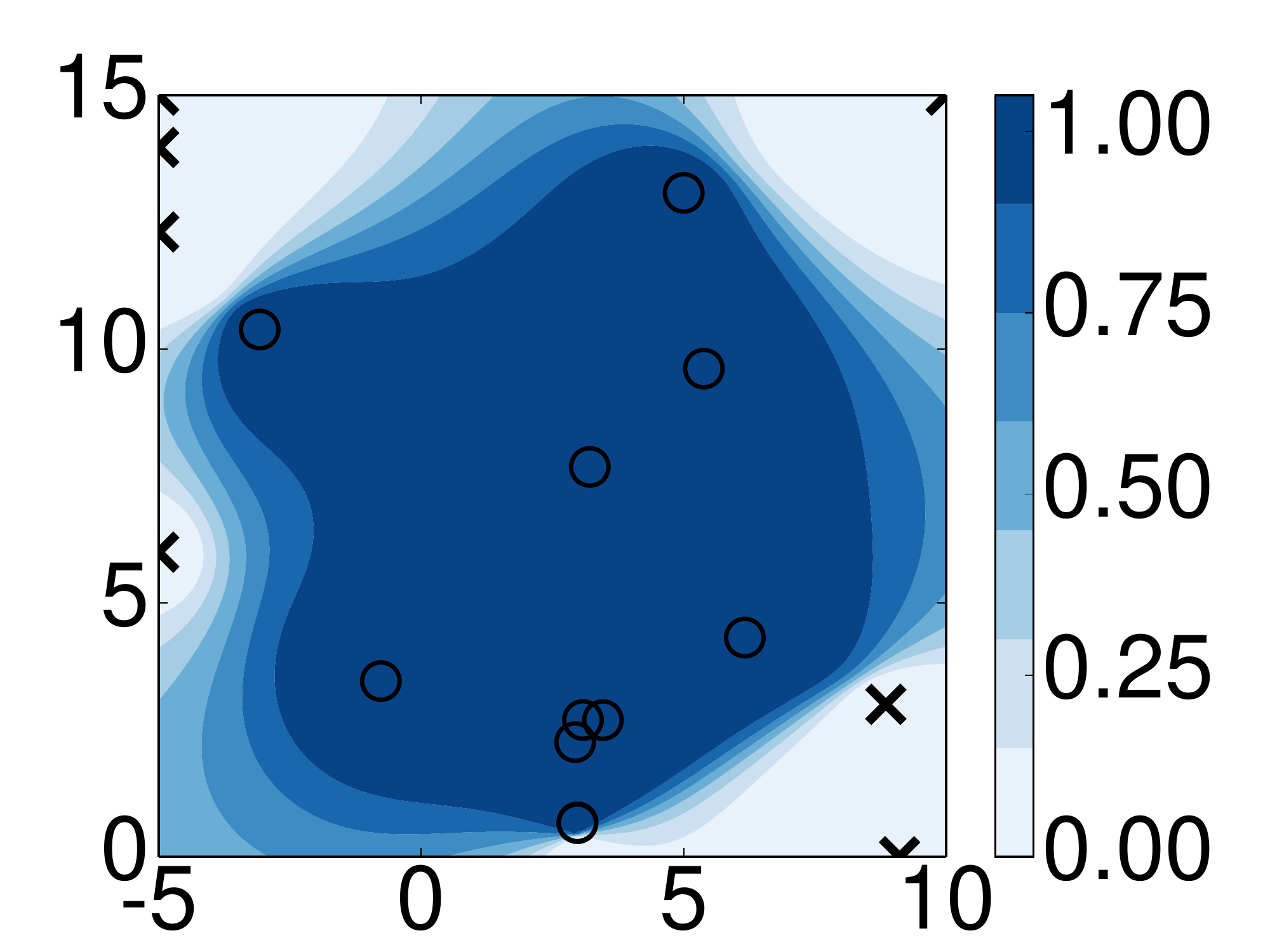}
  \label{fig:circlepv}
}
\subfigure[$\Pr(g(\x)\geq 0) \geq 0.99$]{
  \includegraphics[height=0.17\columnwidth]{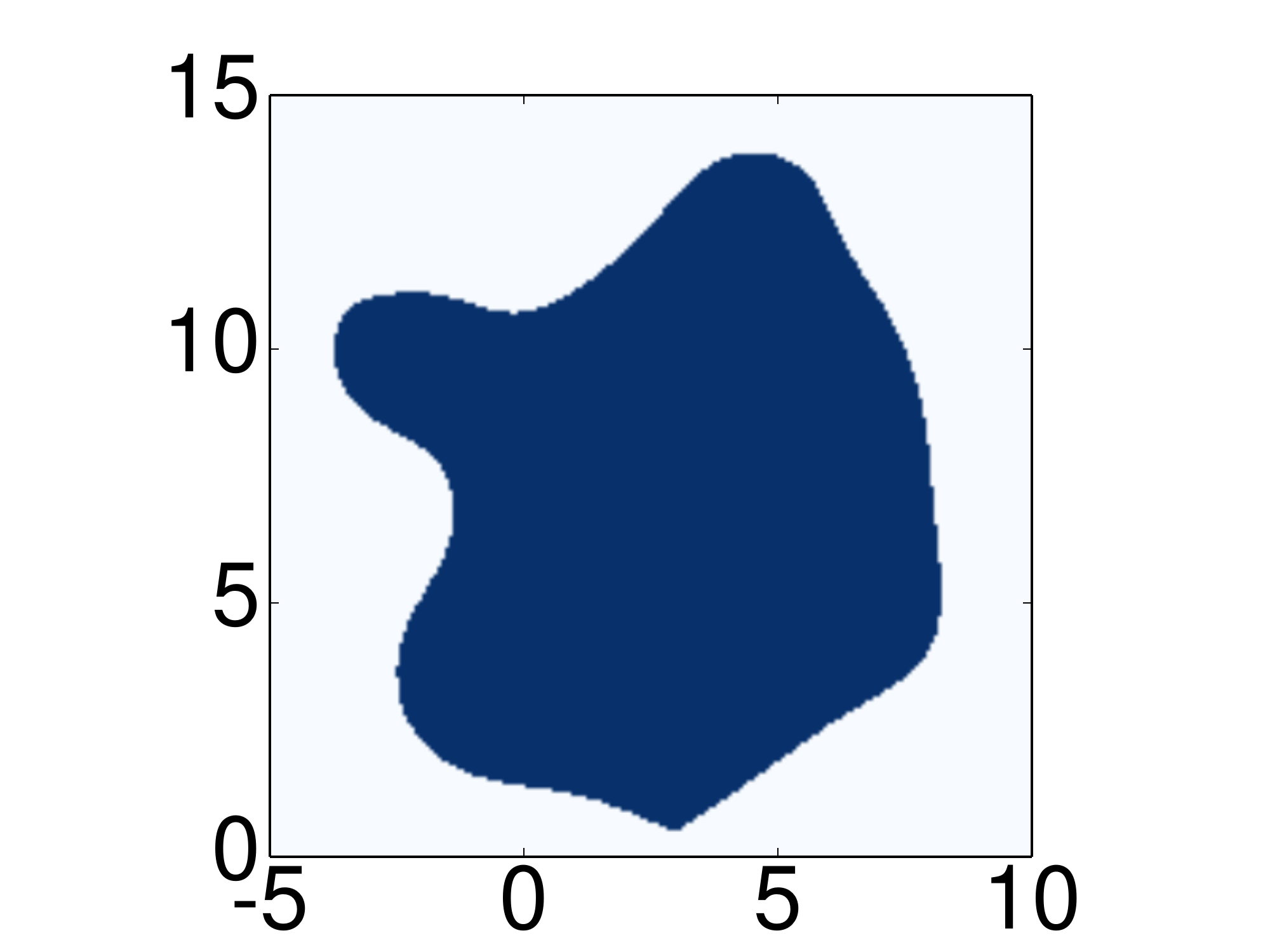}
  \label{fig:circlem}
}
\subfigure[$a(\x)$]{ 
  \includegraphics[height=0.17\columnwidth]{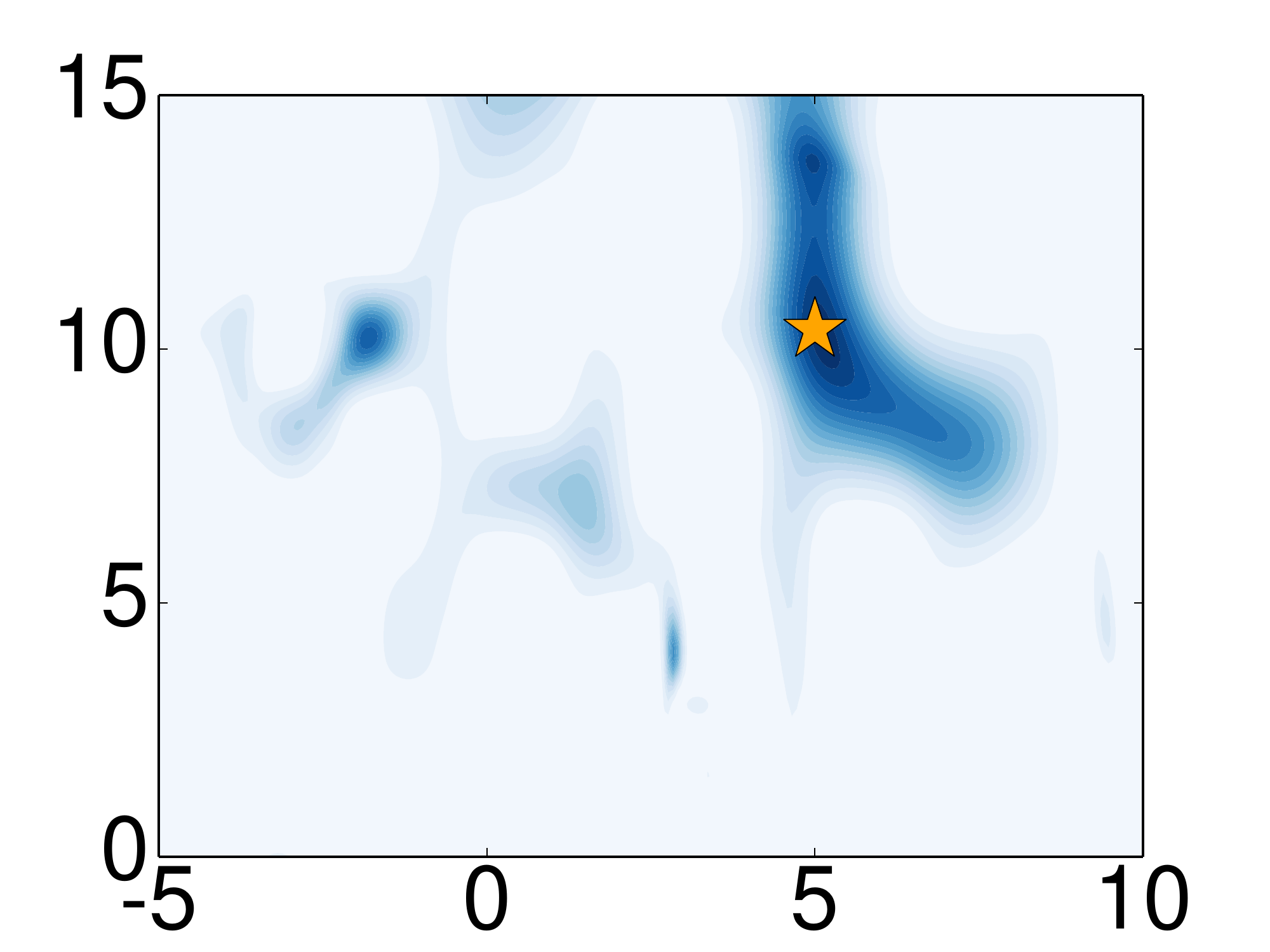}
  \label{fig:circle_acq}
}
\subfigure[$p_\textrm{min}(\x)$]{
  \includegraphics[height=0.17\columnwidth]{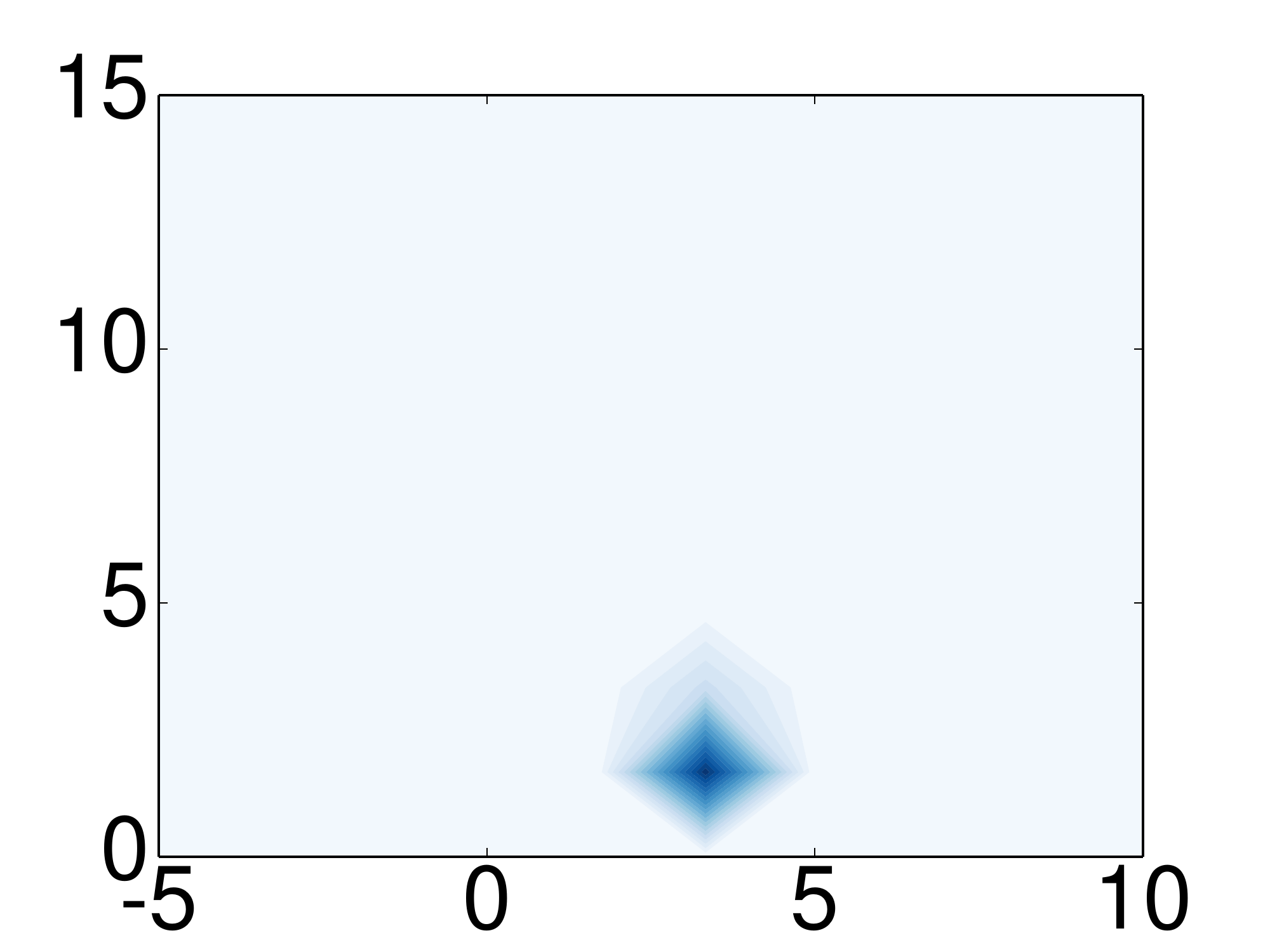}
  \label{fig:circle_pmin}
}

\caption{Constrained Bayesian optimization on the 2D Branin-Hoo function with a disk constraint, after 50 iterations (33 objective evaluations and 17 constraint evaluations): (a) Branin-Hoo function, (b) true constraint, (c) mean of objective function GP, (d) variance of objective function GP, (e) probability of constraint satisfaction, (f) probabilistic constraint,~${\Pr(g(\x)\geq 0)\geq 0.99}$, (g) acquisition function,~$a(\x)$, and (h) probability distribution over the location of the minimum,~$\pmin(\x)$. Lighter colors indicate lower values. Objective function observations are indicated with black circles in (c) and (d). Constraint observations are indicated with black $\times$'s (violations) and o's (satisfactions) in (e). Orange stars: (a) unique true minimum of the constrained problem, (c) best solution 
found by Bayesian optimization, (g) input chosen for the next evaluation, in this case an objective evaluation because $\Delta S_o(\x) > \Delta S_c(\x)$ at the next observation location $\x$.}

\label{fig:branin}
\end{figure*}

\subsection{Acquisition function for decoupled observations}
In some problems, the objective and constraint functions may be evaluated independently. 
We call this property the \emph{decoupling} of the objective and constraint functions. 
In decoupled problems, we must choose to evaluate either the objective function or one of the constraint functions at each iteration of Bayesian optimization. 
As discussed in Section \ref{sec:our_contributions}, it is important to identify problems with this decoupled structure, because often some of the functions are much more expensive to evaluate than others. 
Bayesian optimization with decoupled constraints is a form of multi-task Bayesian optimization \citep{swersky-etal-2013a}, in which the different black-boxes or \emph{tasks} are the objective and decoupled constraint(s), represented by the set $\{\textrm{objective},1,2,\ldots,K\}$ for $K$ constraints.

\subsubsection{Chicken and Egg Pathology}
One possible acquisition function for decoupled constraints is the expected improvement of individually evaluating each task. However, the myopic nature of the EI criterion causes a pathology in this formulation that prevents exploration of the design space. Consider a situation, with a single constraint, in which some feasible region has been identified and thus the current best input is defined, but a large unexplored region remains. Evaluating only the objective in this region could not cause improvement as our belief about~${\Pr(g(\x)\geq 0)}$ will follow the prior and not exceed the threshold ${1-\delta}$. Likewise, evaluating only the constraint would not cause improvement because our belief about the objective will follow the prior and is unlikely to become the new best.  This is a causality dilemma:  we must learn that \emph{both} the objective and the constraint are favorable for improvement to occur, but this is not possible when only a single task is observed.  This difficulty suggests a non-myopic aquisition function which assesses the improvement after a sequence of objective and constraint observations.  However, such a multi-step acquisition function is intractable in general~\citep{ginsbourger-riche-2010}.

Instead, to address this pathology, we propose to use the coupled acquisition function (Eq.~\ref{weighted-ei}) to select an input $\x$ for observation, followed by a second step to determine which task will be evaluated at $\x$. Following \citet{swersky-etal-2013a}, we use the entropy search criterion \citep{hennig-schuler-2012} to select a task. However, our framework does not depend on this choice. 

\subsubsection{Entropy Search Criterion}
Entropy search works by considering $\pmin(\x)$, the probability distribution over the location of the minimum of the objective function. Here, we extend the definition of $\pmin$ to be the probability distribution over the location of the solution to the constrained problem. Entropy search seeks the action that, in expectation, most reduces the relative entropy between~$\pmin(\x)$ and an uninformative base distribution such as the uniform distribution. Intuitively speaking, we want to reduce our uncertainty about $\pmin$ as much as possible at each step, or, in other words, maximize our information gain at each step. Following \citet{hennig-schuler-2012}, we choose $b(\x)$ to be the uniform distribution on the input space. Given this choice, the relative entropy of $\pmin$ and $b$ is the differential entropy of $\pmin$ up to a constant that does not affect the choice of task. Our decision criterion is then
\begin{align}
\label{entropysearch}
T^* = \arg \min_{T} \E_y \left[ S\left(\pmin^{(y_T)}\right) - S(\pmin) \right] \, ,
\end{align}
where $T$ is one of the tasks in $\{\textrm{objective},1,2,\ldots,K\}$, $T^*$ is the selected task,~$S(\cdot)$ is the differential entropy functional, and $\pmin^{(y_T)}$ is $\pmin$ conditioned on observing the value $y_T$ for task $T$. When integrating out the GP covariance hyperparameters, the full form is
\begin{align}
\label{entropysearch-clean-int}
T^* = \arg \min_{T} \int S\left(\pmin^{(y_T)}\right)  \, p\left(y_T|\theta,\omega \right) \mathrm{d}y_T\, \mathrm{d}\theta\, \mathrm{d}\omega
\end{align}
where $y_T$ is a possible observed outcome of selecting task $T$ and $\theta$ and $\omega$ are the objective and constraint GP hyperparameters respectively.\footnote{For brevity, we have omitted the base entropy term (which does not affect the decision $T^*$) and the explicit dependence of $\pmin$ on $\theta$ and $\omega$.}

\subsubsection{Entropy Search in Practice}
Solving Eq. \ref{entropysearch-clean-int} poses several practical difficulties, which we address here in turn. First, estimating~$\pmin(\x)$ requires a discretization of the space. In the spirit of \citet{hennig-schuler-2012}, we form a discretization of $N_d$ points by taking the top $N_d$ points according to the weighted expected improvement criterion. Second, $\pmin$ cannot be computed in closed form and must be either estimated or approximated. \citet{swersky-etal-2013a} use Monte Carlo sampling to estimate $\pmin$ by drawing samples from the GP on the discretization set and finding the minimum. We use the analogous method for constrained optimization: we sample from the objective function GP and all $K$ constraint GPs, and then find the minimum of the objective for which the constraint is satisfied for all $K$ constraint samples.

\subsubsection{Incorporating cost information}
Following \citet{swersky-etal-2013a}, we incorporate information about the relative cost of the tasks by simply scaling the acquisition functions by these costs (provided by the user). In doing so, we pick the task with the most information gain per unit cost. If $\lambda_A$ is the cost of observing task $A$, then Eq. \ref{entropysearch} becomes
\begin{align}
A^* = \arg \min_{A} \frac{1}{\lambda_A} \E_y \left[ S\left(\pmin^{(y_A)}\right) - S(\pmin) \right] \; .
\end{align}

\begin{figure*}[t]
\centering

\subfigure[Online LDA]{
  \includegraphics[trim = 40mm 100mm 40mm 100mm, clip=true, width=0.45\textwidth]{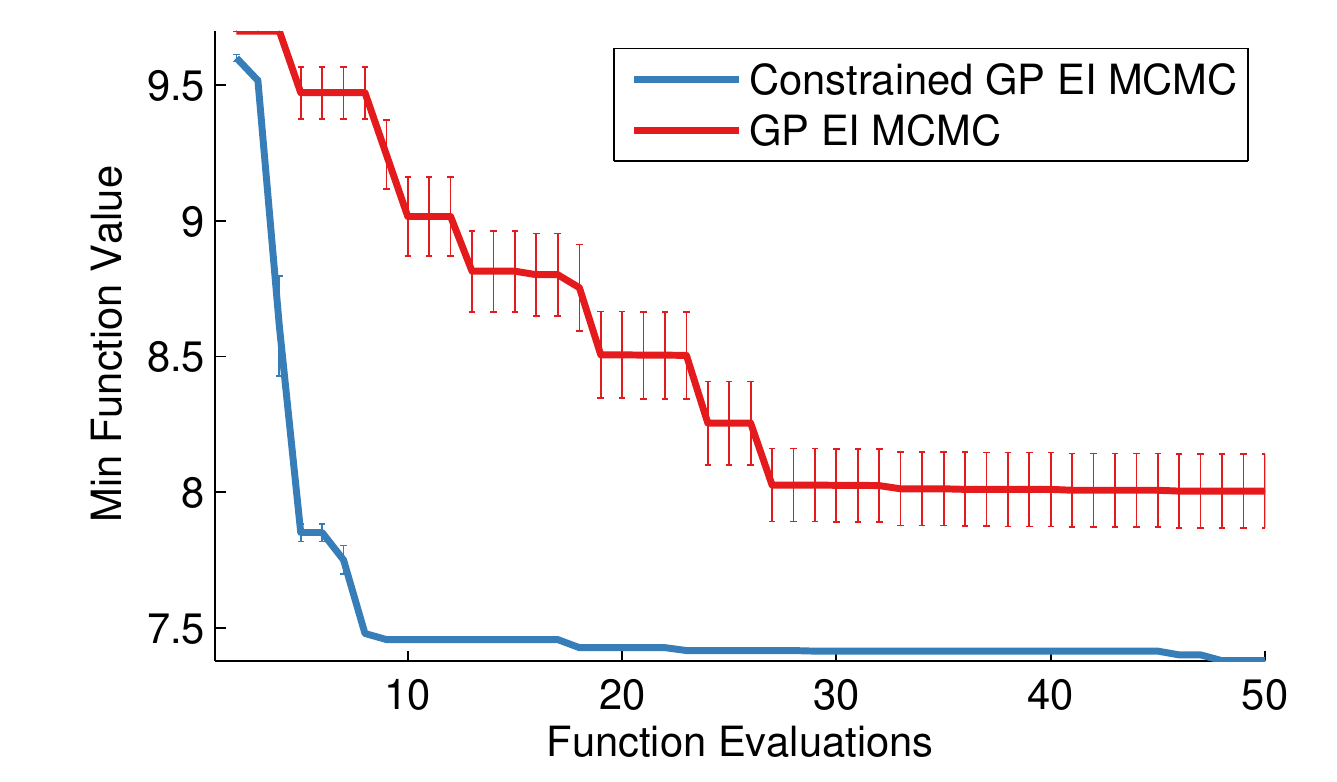}
  \label{fig:online_lda}
}
\subfigure[Neural Network]{
  \includegraphics[trim = 40mm 100mm 40mm 100mm, clip=true, width=0.45\textwidth]{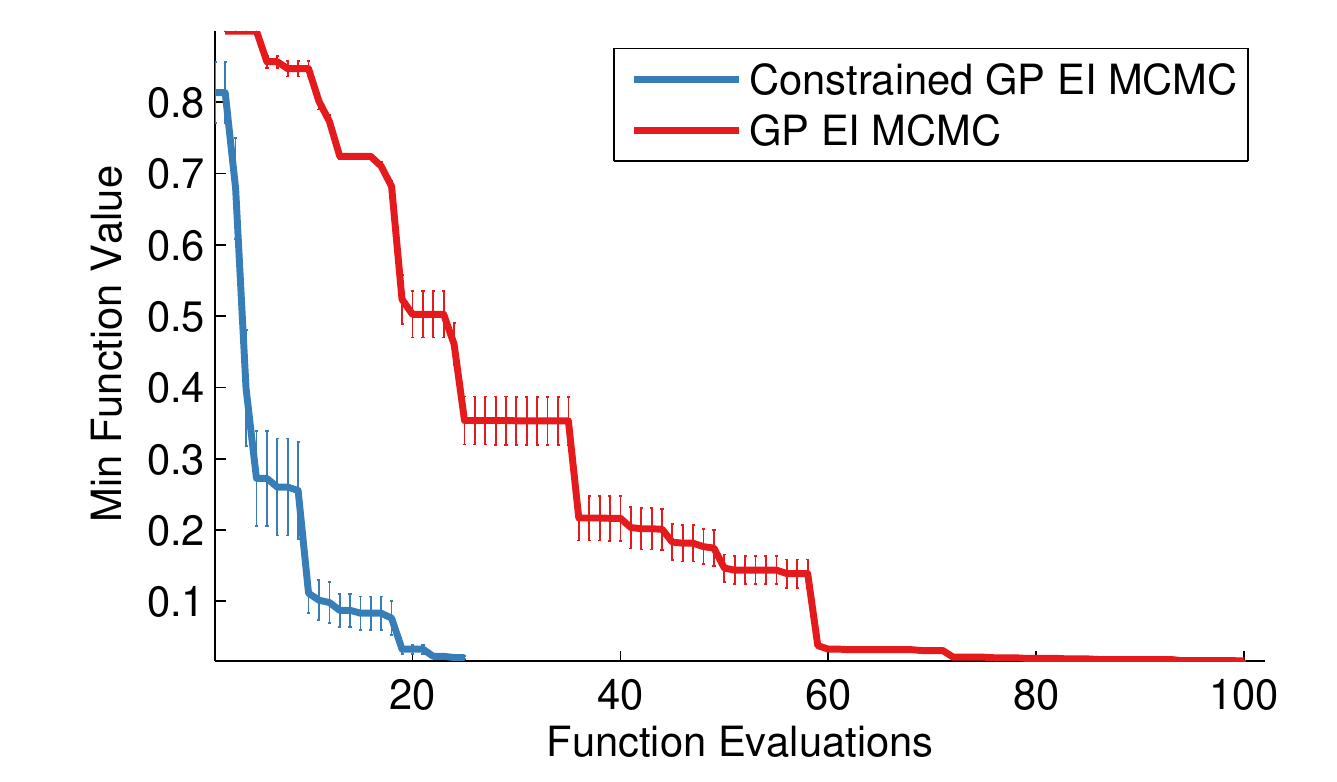}
  \label{fig:neuralnet}
}
\vspace{-0.3em}
\caption{Empirical performance of constrained Bayesian optimization for (a) Online Latent Dirichlet Allocation and (b) turning a deep neural network. Blue curve: our method. Red curve: unconstrained Bayesian optimization with constraint violations as large values. Errors bars indicate standard error from 5 independent runs.}
\label{fig:experimental_results}
\end{figure*}

\begin{figure*}
\centering

\subfigure[Objective Function]{
  \includegraphics[trim = 10mm 0mm 40mm 0mm, clip=true, height=0.18\columnwidth]{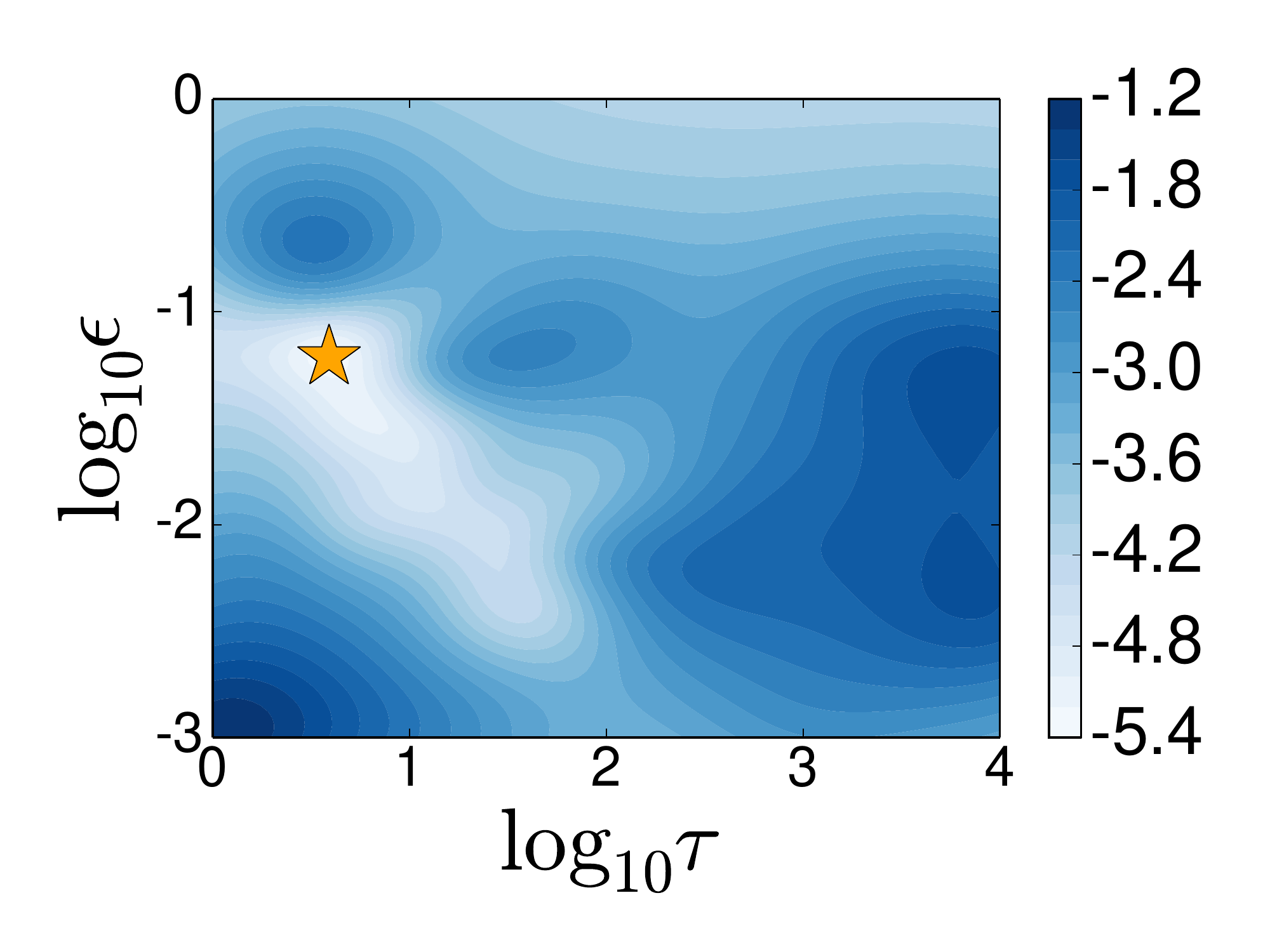}
  \label{fig:mcmc0}
}
\subfigure[Geweke]{
  \includegraphics[trim = 25mm 0mm 40mm 0mm, clip=true, height=0.18\columnwidth]{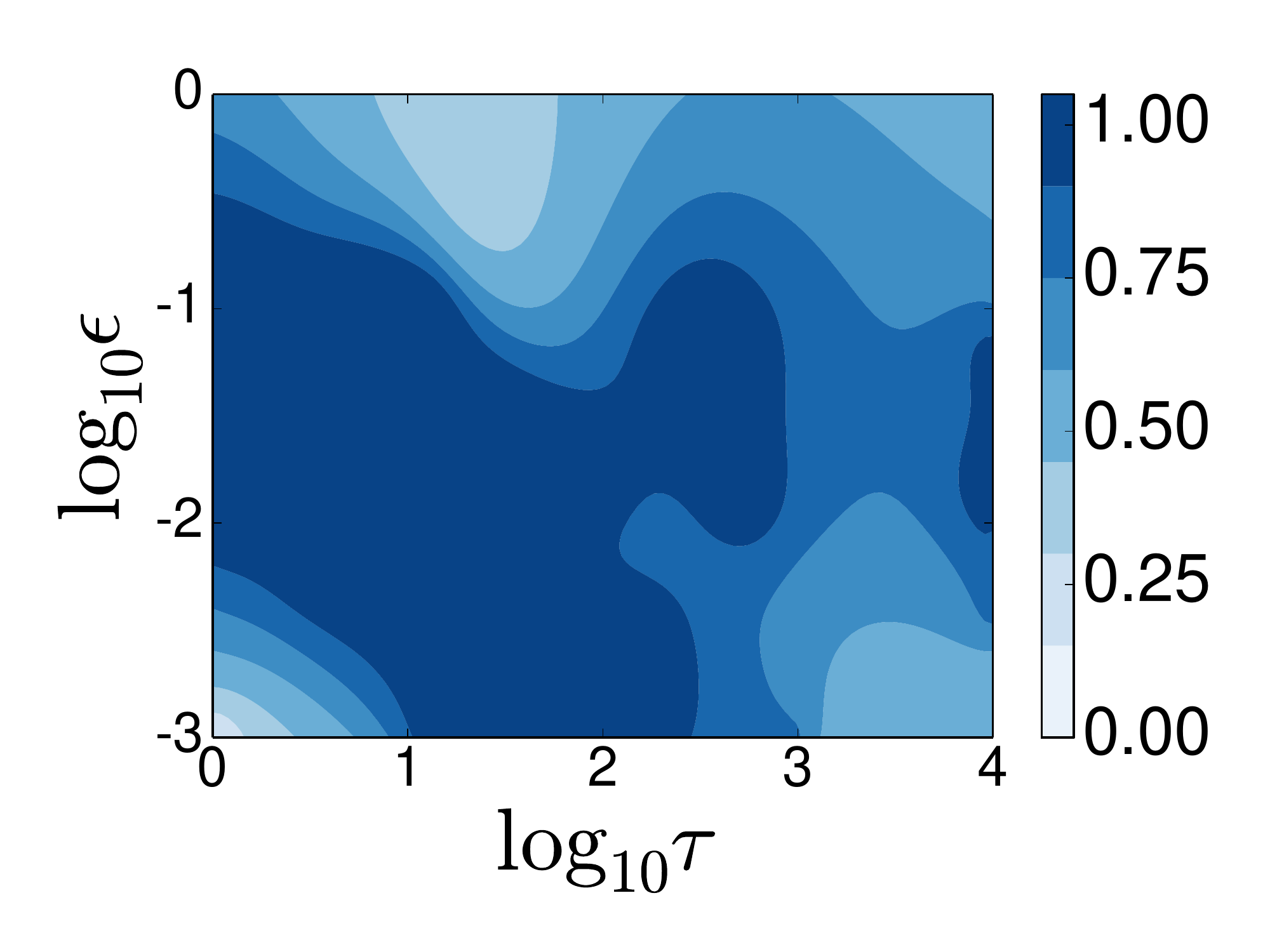}
  \label{fig:mcmc1}
}
\subfigure[Gelman-Rubin]{
  \includegraphics[trim = 25mm 0mm 40mm 0mm, clip=true, height=0.18\columnwidth]{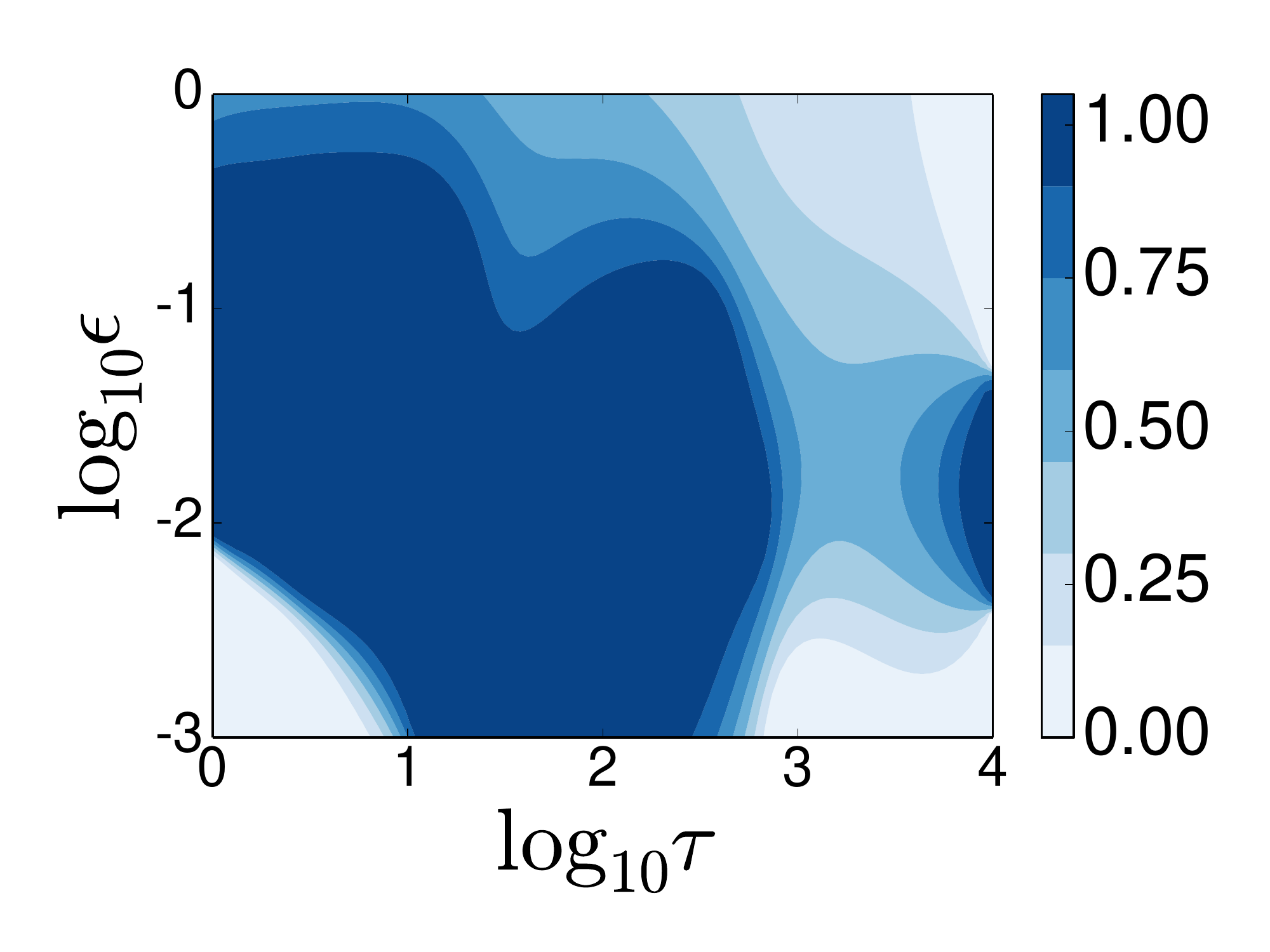}
  \label{fig:mcmc2}
}
\subfigure[Stability]{
  \includegraphics[trim = 25mm 0mm 40mm 0mm, clip=true, height=0.18\columnwidth]{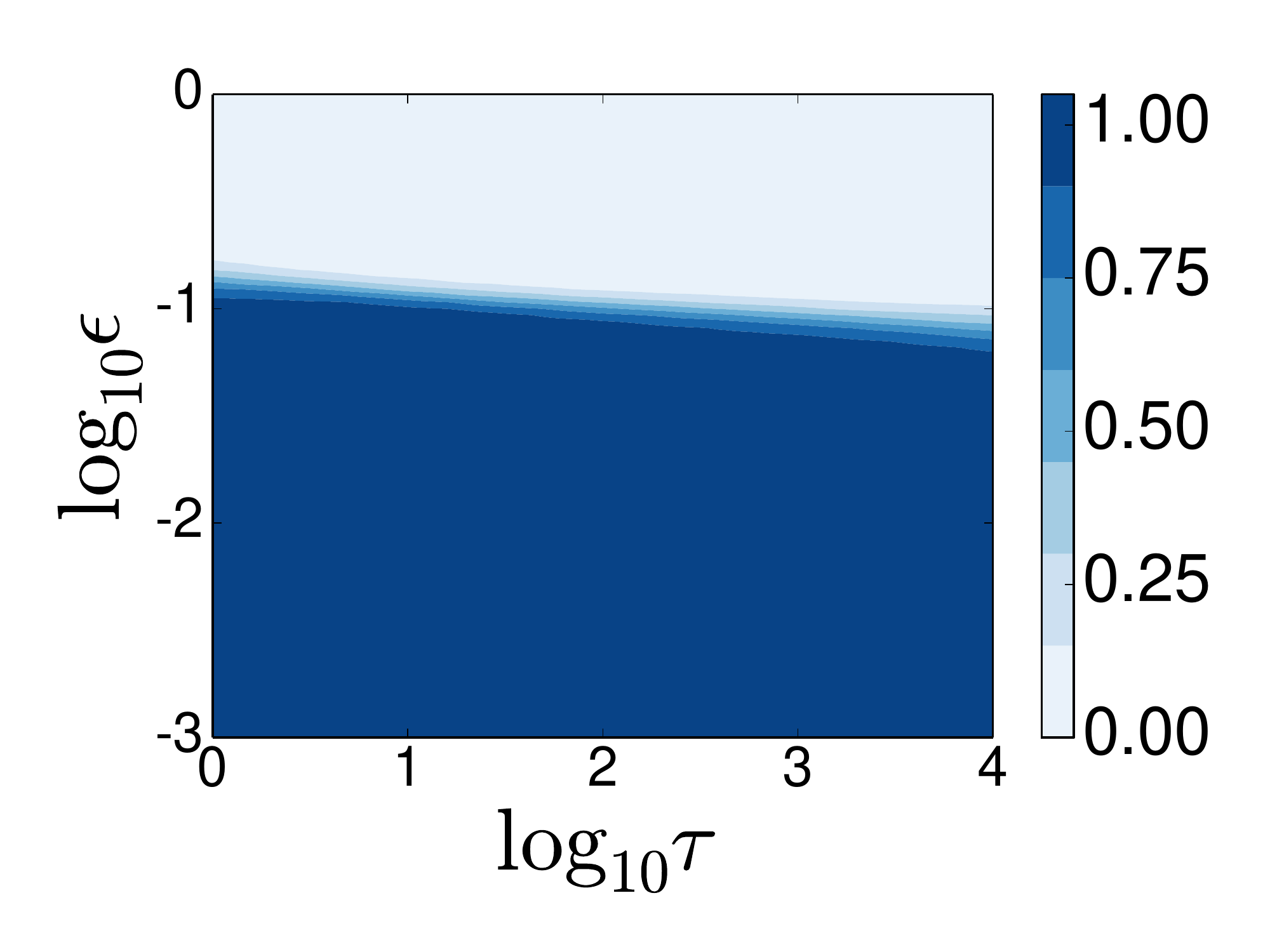}
  \label{fig:mcmc3}
}
\subfigure[Overall]{
  \includegraphics[trim = 25mm 0mm 00mm 0mm, clip=true, height=0.18\columnwidth]{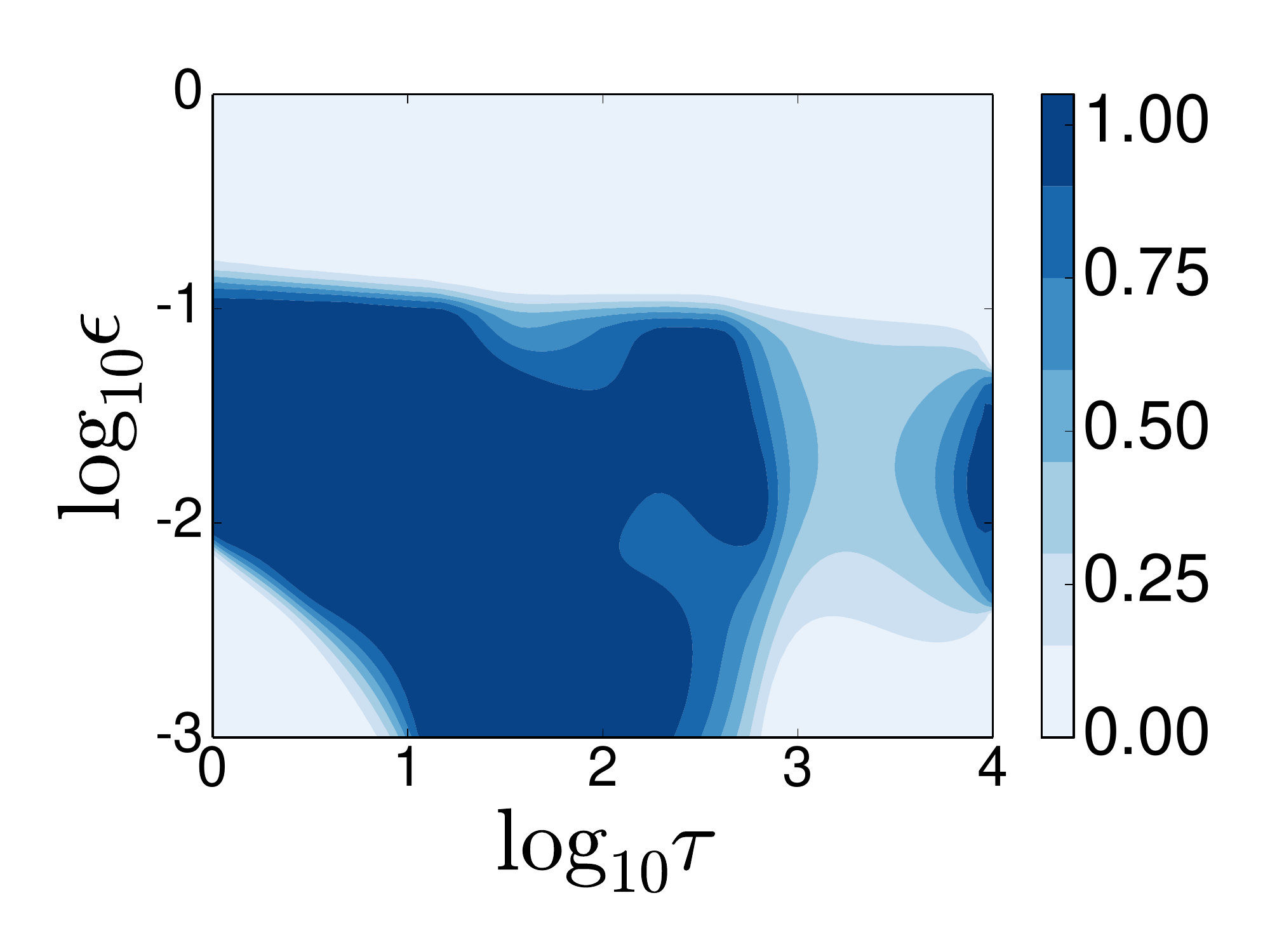}
  \label{fig:mcmc4}
}
\vspace{-0.4em}

\caption{ Tuning Hamiltonian Monte Carlo with constrained Bayesian optimization: (a) objective function model, (b-e) constraint satisfaction probability surfaces for (b) Geweke test, (c) Gelman-Rubin test, (d) stability of the numerical integration, (d) overall, which is the product of the preceding three probability surfaces. In (a), lighter colors correspond to more effective samples, circles indicate function evaluations, and the orange star indicates the best solution. In (b-e), constraint observations are indicated with black $\times$'s (violations) and o's (satisfactions). Vertical axis label at left is for all subplots. Probability colormap at right is for (b-d).}

\label{fig:mcmc}
\end{figure*}

\section{Experiments}
\label{section:results}

\subsection{Branin-Hoo function}
We first illustrate constrained Bayesian optimization on the Branin-Hoo function, a 2D function with three global minima (Fig.~\ref{fig:truebranin}). We add a decoupled disk constraint ${(\x_1 - 2.5)^2 + (\x_2 - 7.5)^2) \leq 50}$, shown in Fig.~\ref{fig:diskcon}. This constraint eliminates the upper-left and lower-right solutions, leaving a unique global minimum at ${\x=(\pi, 2.275)}$, indicated by the orange star in Fig.~\ref{fig:truebranin}. 
After 33 objective function evaluations and 17 constraint evaluations, the best solution is
$(3.01, 2.36)$, which satisfies the constraint and has value 0.48 (true best value = 0.40).

\subsection{Online LDA with sparse topics}
Online Latent Dirichlet Allocation \citep[LDA,][]{Hoffman2010} is an efficient variational formulation of a popular topic model for learning topics and corresponding word distributions given a corpus of documents. In order for topics to have meaningful semantic interpretations, it is desirable for the word distributions to exhibit sparsity. In this experiment we optimize the hyperparameters of online LDA subject to the constraint that the entropy of the per-topic word distribution averaged over topics is less than $\log_2 200$ bits, 
which is achieved, for example by allocating uniform density over 200 words. We used the online LDA implementation from \citet{agarwal-etal-2011a} and optimized five hyperparameters corresponding to the number of topics (from 2~to~100), two Dirichlet distribution prior base measures (from 0~to~2), and two learning rate parameters (rate from 0.1~to~1, decay from $10^{-5}$~to~1).  As a baseline, we compare with unconstrained Bayesian optimization in which constraint violations are set to the worst possible value for this LDA problem.
Fig.~\ref{fig:online_lda} shows that constrained Bayesian optimization significantly outperforms the baseline. Intuitively, the baseline is poor because the GP has difficulty modeling the sharp discontinuities caused by the large values. 
\begin{table*}[htbp]
\caption{Tuning Hamiltonian Monte Carlo.}
\label{table:mcmc}
\begin{center}
\begin{tabular}{llllllll}
\multicolumn{1}{c}{Experiment} & \multicolumn{1}{c}{burn-in} &\multicolumn{1}{c}{$\tau$}  &\multicolumn{1}{c}{$\epsilon$}  & \multicolumn{1}{c}{mass}  & \multicolumn{1}{c}{\# samples}  &\multicolumn{1}{c}{accept}  &\multicolumn{1}{c}{\bf effective samples} \\
\hline \\
Baseline & 10\%  & 100 & 0.047   & 1    & $8.3\times 10^3$     & 85\%  & $1.1\times10^3$   \\  
BayesOpt & 3.8\% & 2   & 0.048 & 1.55 & $3.3\times 10^5$	 & 70\% & $9.7\times 10^4$   
\end{tabular}
\end{center}
\vspace{-1em}
\end{table*}

\subsection{Memory-limited neural net}
In the final experiment, we optimize the hyperparameters of a deep neural network on the MNIST handwritten digit classification task in a memory-constrained scenario.  We optimize over 11 parameters: 1 learning rate, 2 momentum parameters (initial and final), the number of hidden units per layer (2 layers), the maximum norm on model weights (for 3 sets of weights), and the dropout regularization probabilities (for the inputs and 2 hidden layers). We optimize the classification error on a withheld validation set under the constraint that the total number of model parameters (weights in the network) must be less than one million.  This constraint is decoupled from the objective and inexpensive to evaluate, because the number of weights can be calculated directly from the parameters, without training the network.  We train the neural network using momentum-based stochastic gradient descent which is notoriously difficult to tune as training can diverge under various combinations of the momentum and learning rate. 
When training diverges, the objective function cannot be measured. Reporting the constraint violation as a large objective value performs poorly because it introduces sharp discontinuities that are hard to model (Fig. \ref{fig:experimental_results}). 
This necessitates a second noisy, binary constraint which is violated when training diverges, for example when the both the learning rate and momentum are too large.   The network is trained\footnote{We use the Deepnet package: \url{https://github.com/nitishsrivastava/deepnet}.} for 25,000 weight updates and the objective is reported as classification error on the standard validation set.  Our Bayesian optimization routine can thus choose between two decoupled tasks, evaluating the memory constraint or the validation error after a full training run.  Evaluating the validation error can still cause a constraint violation when the training diverges, which is treated as a binary constraint in our model.  Fig. \ref{fig:neuralnet} shows a comparison of our constrained Bayesian optimization against a baseline standard Bayesian optimization where constraint violations are treated as resulting in a random classifier (90\% error).  Only the objective evaluations are presented, since constraint evaluations are extremely inexpensive compared to an entire training run.  In the event that training diverges on an objective evaluation, we report 90\% error.  The optimized net has a learning rate of 0.1, dropout probabilities of 0.17 (inputs), 0.30 (first layer), and 0 (second layer), initial momentum 0.86, and final momentum 0.81. Interestingly, the optimization chooses a small first layer (size 312) and a large second layer (size 1772).

\subsection{Tuning Markov chain Monte Carlo}

Hamiltonian Monte Carlo (HMC) is a popular MCMC sampling technique that takes advantage of gradient information for rapid mixing. However, HMC contains several parameters that require careful tuning. The two basic parameters are the number of leapfrog steps $\tau$, and the  step size $\epsilon$. HMC may also include a mass matrix which introduces $\mathcal{O}(D^2)$ additional parameters in $D$ dimensions, although the matrix is often chosen to be diagonal ($D$ parameters) or a multiple of the identity matrix (1 parameter) \citep{nealbook}. In this experiment, we optimize the performance of HMC using Bayesian optimization; see \citet{mahendran-2012a} for a similar approach. We optimize the following parameters: $\tau$, $\epsilon$, a mass parameter, and the fraction of the allotted computation time spent burning in the chain.

Our experiment measures the number of effective samples (ES) in a fixed computation time; this corresponds to finding chains that minimize estimator variance. We impose the constraints that the generated samples must pass the Geweke \citep{Geweke92evaluatingthe} and Gelman-Rubin \citep{gelmanrubin} convergence diagnostics. In particular, we require the worst (largest absolute value) Geweke test score across all variables and chains to be at most 2.0, and the worst (largest) Gelman-Rubin score between chains and across all variables to be at most 1.2. We use PyMC \citep{PyMC} for the convergence diagnostics and the LaplacesDemon R package to compute effective sample size. The chosen thresholds for the convergence diagnostics are based on the PyMC and LaplacesDemon documentation. The HMC integration may also diverge for large values of $\epsilon$; we treat this as an additional constraint, and set ${\delta=0.05}$ for all constraints. 
We optimize HMC sampling from the posterior of a logistic regression binary classification problem using the German credit data set from the UCI repository \citep{Frank2010}. The data set contains 1000 data points, and is normalized to have unit variance. 
We initialize each chain randomly with $D$ independent draws from a Gaussian distribution with mean zero and standard deviation $10^{-3}$.
For each set of inputs, we compute two chains, each with 5 minutes of computation time on a single core of a compute node.

Fig. \ref{fig:mcmc} shows the constraint surfaces discovered by Bayesian optimization for a simpler experiment in which only $\tau$ and $\epsilon$ are varied; burn-in is fixed at 10\% and the mass is fixed at 1. These diagrams yield interpretations of the feasible region; for example, Fig. \ref{fig:mcmc3} shows that the numerical integration diverges for values of $\epsilon$ above ${\approx 10^{-1}}$. Table \ref{table:mcmc} shows the results of our 4-parameter optimization after 50 iterations, compared with a baseline that is reflective of a typical HMC configuration: 10\% burn in, 100 leapfrog steps, and the step size chosen to yield an 85\% proposal accept rate. Each row in the table was produced by averaging 5 independent runs with the given parameters. The optimization chooses to perform very few (${\tau=2}$) leapfrog steps and spend relatively little time (3.8\%) burning in the chain, and chooses an acceptance rate of 70\%. In contrast, the baseline spends much more time generating each proposal (${\tau=100}$), which produces many fewer total samples and, correspondingly, significantly fewer effective samples.

\section{Conclusion}
\label{section:conclusion}
In this paper we extended Bayesian optimization to constrained optimization problems. Because constraint observations may be noisy, we formulate the problem using probabilistic constraints, allowing the user to directly express the tradeoff between cost and risk by specifying the confidence parameter $\delta$. We then propose an acquisition function to perform constrained Bayesian optimization, including the case where the objective and constraint(s) may be observed independently. We demonstrate the effectiveness of our system on the meta-optimization of machine learning algorithms and sampling techniques. Constrained optimization is a ubiquitous problem and we believe this work has applications in areas such as product design (e.g. designing a low-calorie cookie), machine learning meta-optimization (as in our experiments), real-time systems (such as a speech recognition system on a mobile device with speed, memory, and/or energy usage constraints), or any optimization problem in which the objective function and/or constraints are expensive to evaluate and possibly noisy.

\subsubsection*{Acknowledgements}
The authors would like to thank Geoffrey Hinton, George Dahl, and Oren Rippel for helpful discussions, and Robert Nishihara for help with the experiments. This work was partially funded by DARPA Young Faculty Award N66001-12-1-4219. Jasper Snoek is a fellow in the Harvard Center for Research on Computation and Society.

\bibliography{constrained}
\bibliographystyle{plainnat}

\end{document}